\documentclass[a4paper,12pt,onecolumn]{article}

\usepackage{FAS}
\usepackage[latin1]{inputenc}
\usepackage{graphicx}
\usepackage{xcolor}
\usepackage{float}
\usepackage{gensymb}
\usepackage{caption} 
\begin{document}
\title{Understanding the Domain Gap in LiDAR Object Detection Networks}
\author{Jasmine Richter \and Florian Faion \and Di Feng \and Paul Benedikt Becker \and Piotr Sielecki \and Claudius Gläser %
\thanks{Robert Bosch GmbH (e-mail: firstname.lastname@bosch.com).}}
\date{}
\maketitle \thispagestyle{empty}

\begin{abstract}
In order to make autonomous driving a reality, artificial neural networks have to work reliably in the open-world. However, the open-world is vast and continuously changing, so it is not technically feasible to collect and annotate training datasets which accurately represent this domain. Therefore, there are always domain gaps between training datasets and the open-world which must be understood. In this work, we investigate the domain gaps between high-resolution and low-resolution LiDAR sensors in object detection networks. Using a unique dataset, which enables us to study sensor resolution domain gaps independent of other effects, we show two distinct domain gaps - an inference domain gap and a training domain gap. The inference domain gap is characterised by a strong dependence on the number of LiDAR points per object, while the training gap shows no such dependence. These findings show that different approaches are required to close these inference and training domain gaps.
\end{abstract}


\section{Introduction}
The open-world in which we live is diverse and constantly changing, so autonomous vehicles have to work robustly in frequently changing weather conditions \cite{hnewa_rain_2021}, light conditions \cite{singha_night_2019}, and in different locations such as cities, highways and rural areas. In addition to this, the hardware used in autonomous vehicles is continuously being improved and updated. This trend is clearly observable in LiDAR sensors, which are continually being manufactured with higher and higher resolutions \cite{wang_lidar_2020}. Collecting and labelling datasets covering the long-tail of open-world scenarios for every new sensor generation is prohibitively time-consuming and expensive \cite{li_data_2019}. Therefore, it is important to understand the domain gaps between datasets recorded with LiDARs of different resolutions - the sensor-to-sensor domain gaps - so that these domain gaps can be closed and datasets and models can be reused and improved.

Although domain adaptation has been widely studied \cite{ganin_unsupervised_2015, triess_survey_2021, wirges_single-stage_2020, rist_cross_sensor_2019, qin_multiscale_2019, xu_spg_2021}, to the best of our knowledge, this is the first time that the distinct inference and training domain gaps have been identified and investigated in LiDAR object detection networks. In this work, the inference domain gap refers to the domain gap observed when the network is trained with data from a particular LiDAR and then evaluated on data from the same LiDAR and a different LiDAR. Whereas the training domain gap refers to the domain gap observed when two networks are trained, the first with data from one LiDAR and the second with data from another LiDAR, and then both evaluated on data from one of those LiDARs. This distinction is only identifiable because, unlike other autonomous driving datasets \cite{geiger_kitti_2013, caesar_nuscenes_2020}, our dataset enables us to study the sensor-to-sensor domain gaps directly and independently of other common domain gaps, such as from different weather conditions, light conditions and locations.

\section{Dataset}
In order to isolate the sensor-to-sensor domain gaps from other domain gaps, we used a unique dataset in which LiDAR data from high-resolution Hesai Pandar64P sensors and low-resolution Velodyne VLP-32C sensors was recorded simultaneously. This dataset was collected on a drive of 7,800 km, visiting 33 cities in 6 European countries. The route was chosen to maximise scenario variety and includes highway, rural and urban driving in diverse environmental conditions including day and night, as well as clear and adverse weather. For this work, we used around 13,100 single frames containing around 432,000 labelled bounding boxes in total.

\begin{figure}[H]
	\centering
	\includegraphics[width=14cm]{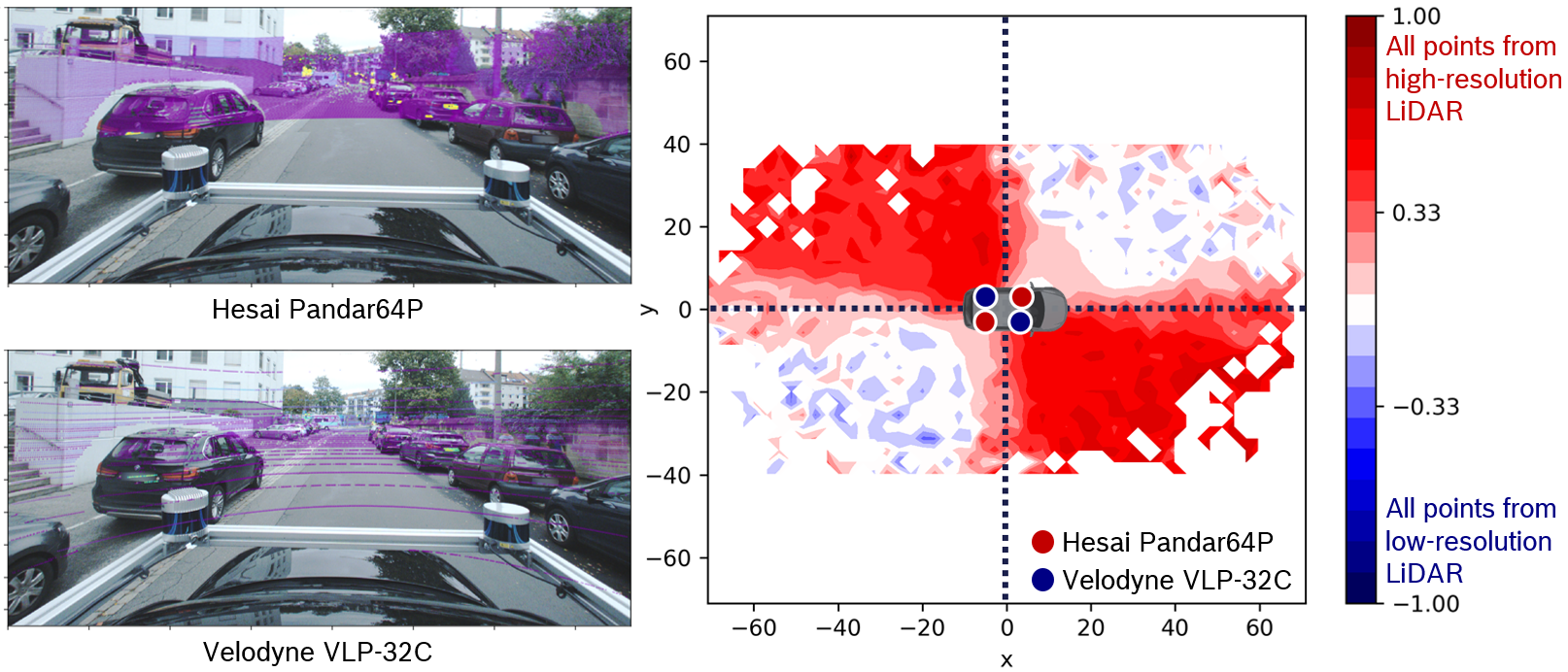}
	\caption{Configuration of the LiDAR sensors. Left: LiDAR points overlayed on camera images for the high-resolution Hesai Pandar64P sensors and low-resolution Velodyne VLP-32C sensors respectively. Right: Relative number of points received per object from the high-resolution Hesai Pandar64P sensors and low-resolution Velodyne VLP-32C sensors as a function of (x, y) location in the bird's eye view (BEV).} \label{fig:camera_and_points_map}
\end{figure}

The high-resolution and low-resolution LiDAR sensors were mounted in an alternating pattern on the roof of the car, as shown in Figure~\ref{fig:camera_and_points_map}. These alternating mounting positions ensure that both the high-resolution and low-resolution sensors cover the full 360\degree\ field-of-view. The different sensor resolutions and fields-of-view cause an uneven distribution of points to be received from the objects in the dataset, with slightly more points received from the low-resolution sensors in the bottom-left and top-right of the BEV (where points are received from two low-resolution sensors and one high-resolution sensor) and significantly more points received from the high-resolution sensors in the top-left and bottom-right of the BEV (where points are received from two high-resolution sensors and one low-resolution sensor).

\section{Experiments}
Our experiments were carried out on an object detection network based on the PIXOR architecture \cite{Yang_2018_CVPR}. Identical randomly initialised weights and hyperparameters were used for all trainings. First, we quantitatively compared relative average precisions. Then we explored how relative average precision varies spatially over the BEV. Finally, we investigated the relationship between recall and the average number of points received per object. 

\subsection{Relative average precision}
In order to quantify the distinct inference and training sensor-to-sensor domain gaps, we trained and evaluated using all sensor combinations. The relative average precisions (APs) of these trainings and evaluations are shown in Table~\ref{table:all}. The relative difference in the APs indicate a fundamental dissimilarity between the inference domain gap and the training domain gap. 

\begin{table}[H]
	\renewcommand{\arraystretch}{1.3}
	\caption{Relative APs of object detection networks trained and evaluated on different combinations of high-resolution and low-resolution LiDAR sensors. Higher is better. Top: Inference domain gap. Bottom: Training domain gap.} \label{table:all} \centering
	\begin{tabular}{|c|c||c|}
		\hline
		Trained on & Evaluated on & Relative AP \\
		\hline
		\hline
		High-res & High-res & \textbf{+25\%} \\
		\hline
		High-res & Low-res & 0\% \\
		\hline
		\noalign{\vskip 0.4cm} 
	\end{tabular}
	\begin{tabular}{|c|c||c|}
		\hline
		Trained on & Evaluated on & Relative AP \\
		\hline
		\hline
		Low-res & High-res & \textbf{+11\%} \\
		\hline
		Low-res & Low-res & 0\% \\
		\hline
		\noalign{\vskip 0.4cm} 
	\end{tabular}
	\begin{tabular}{|c|c||c|}
		\hline
		Trained on & Evaluated on & Relative AP \\
		\hline
		\hline
		High-res & High-res & \textbf{+7\%} \\
		\hline
		Low-res & High-res & 0\% \\
		\hline
	\end{tabular}
	\begin{tabular}{|c|c||c|}
		\hline
		Trained on & Evaluated on & Relative AP \\
		\hline
		\hline
		High-res & Low-res & 0\% \\
		\hline
		Low-res & Low-res & \textbf{+6\%} \\
		\hline
	\end{tabular}
\end{table}

\subsubsection{Inference domain gap}
For the inference domain gap (trained on high-resolution and evaluated on high-resolution and low-resolution, or trained on low-resolution and evaluated on high-resolution and low-resolution), the relative AP is higher when the evaluation is carried out using the high-resolution LiDAR data, independent of whether the object detection network was trained with high-resolution or low-resolution LiDAR data. This shows that using a higher-resolution LiDAR sensor at inference time can improve the performance of the object detection network, even if the network was trained with data from a different LiDAR sensor.

\subsubsection{Training domain gap}
In contrast to the inference domain gap, for the training domain gap (trained on high-resolution and low-resolution, then evaluated on either high-resolution or low-resolution), the relative AP is higher when the evaluation is carried out using LiDAR data from the same sensor that was used for training. In this case, there is no evidence that using a higher-resolution LiDAR sensor at training time leads to any improvement in the object detection network performance. In fact, it actually results in worse performance. The object detection network performs best when it is trained using data from the same LiDAR sensor that is used for the evaluation.

\subsection{Spatial variation of relative average precision}
\begin{figure}[H]
	\centering
	\includegraphics[width=16cm]{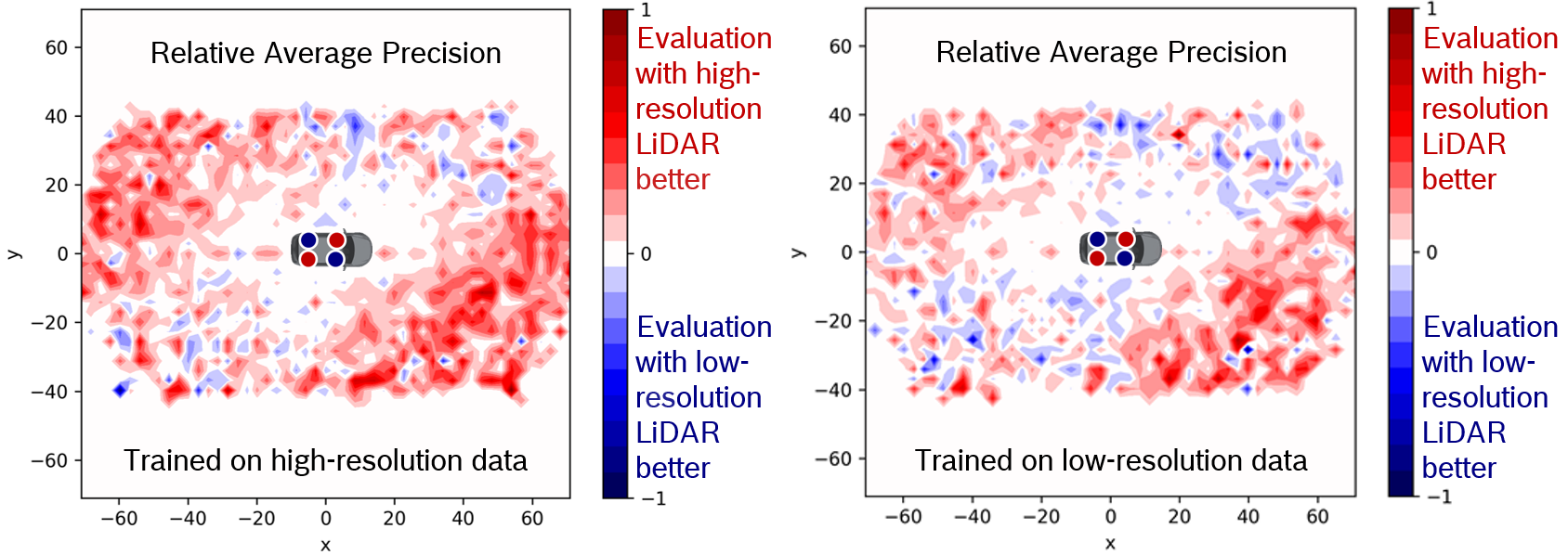}
	\caption{Relative AP of all objects as a function of (x, y) location in the BEV. Left: Trained on high-resolution LiDAR data. Right: Trained on low-resolution LiDAR data.} \label{fig:train_same}
\end{figure}

\begin{figure}[H]
	\centering
	\includegraphics[width=16cm]{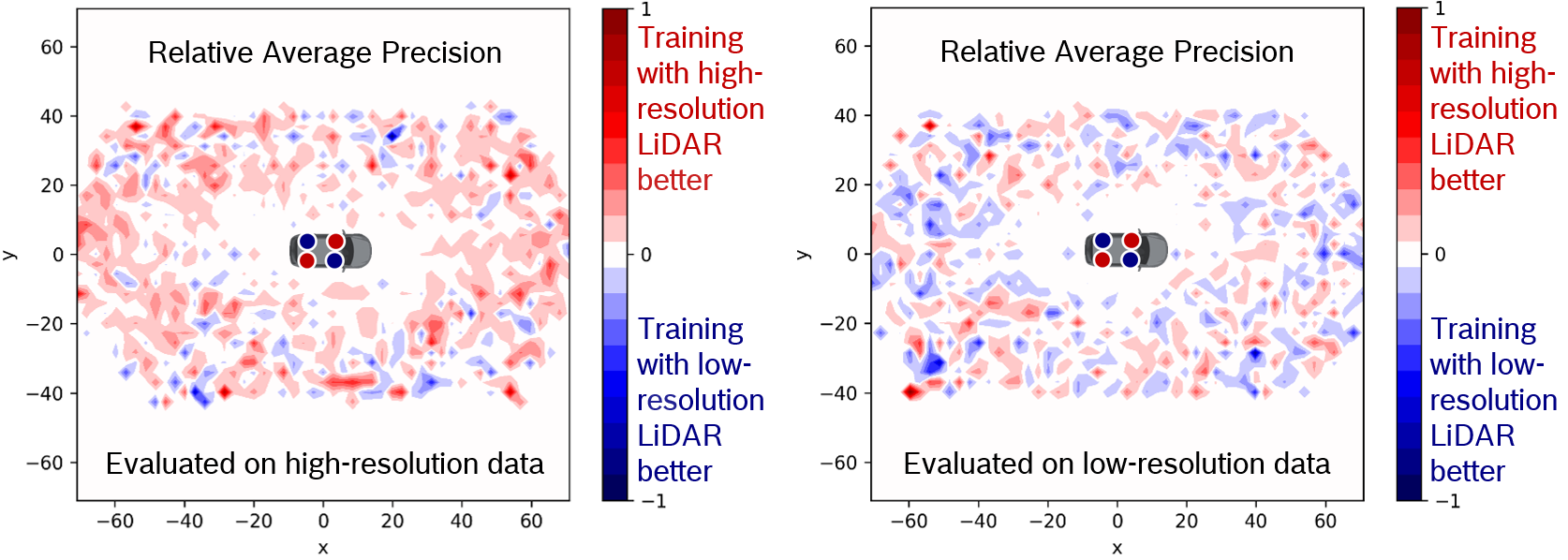}
	\caption{Relative AP of all objects as a function of (x, y) location in the BEV. Left: Evaluated on high-resolution LiDAR data. Right: Evaluated on low-resolution LiDAR data.} \label{fig:eval_same}
\end{figure}

In order to further understand the domain gaps observed in the relative APs, we explored how relative AP varies spatially over the BEV. Figure~\ref{fig:train_same} shows the relative AP of all objects as a function of (x, y) location in the BEV, for the networks trained on high-resolution and low-resolution LiDAR data respectively. While Figure~\ref{fig:eval_same} shows the relative AP of all objects as a function of (x, y) location in the BEV, for the networks evaluated on high-resolution and low-resolution LiDAR data respectively.

\subsubsection{Inference domain gap}
 For the inference domain gap, independent of whether the network was trained on high-resolution or low-resolution LiDAR data, the relative AP is not evenly distributed over the entire BEV. Instead, for both plots in Figure~\ref{fig:train_same}, the relative AP is higher in the top-left and bottom-right of the BEV. This distribution of relative AP looks very similar to the distribution of relative points per object shown in Figure~\ref{fig:camera_and_points_map}. Again, this indicates that no matter which LiDAR data is used for training, the performance of the object detection network can be improved at inference time by using a higher-resolution LiDAR sensor at inference time.

\subsubsection{Training domain gap}
In contrast to the inference domain gap, for the training domain gap, the relative AP is evenly distributed over the entire BEV for both plots in Figure~\ref{fig:eval_same} and there are no specific regions where the relative AP is higher or lower. Instead, when the networks trained on high-resolution and low-resolution data are evaluated on high-resolution data, the relative AP is generally higher for the object detection network trained on high-resolution data. The same is true when the network trained on low-resolution data is evaluated on low-resolution data. Again, this indicates that using a higher-resolution LiDAR at training time does not improve the performance of the object detection network. Instead, training with data from the same LiDAR sensor that is used for evaluation leads to the best object detection network performance.

\subsection{Relationship between recall and average number of points received per object}
Since both the relative APs and spatial variation of relative APs indicate a relationship between the number of points received from an object and the performance of the object detection network, we further explored this relationship by investigating how the recall varies with the average number of points received per object. Using recall, which is defined as the proportion of true positives discovered from the labelled real-world objects, enables us to understand how many points on average are required to correctly identify real-world objects. Such a metric is important for autonomous driving, because failing to identify real-world objects could pose a significant safety risk.

\begin{figure}[H]
	\centering
	\includegraphics[width=16cm]{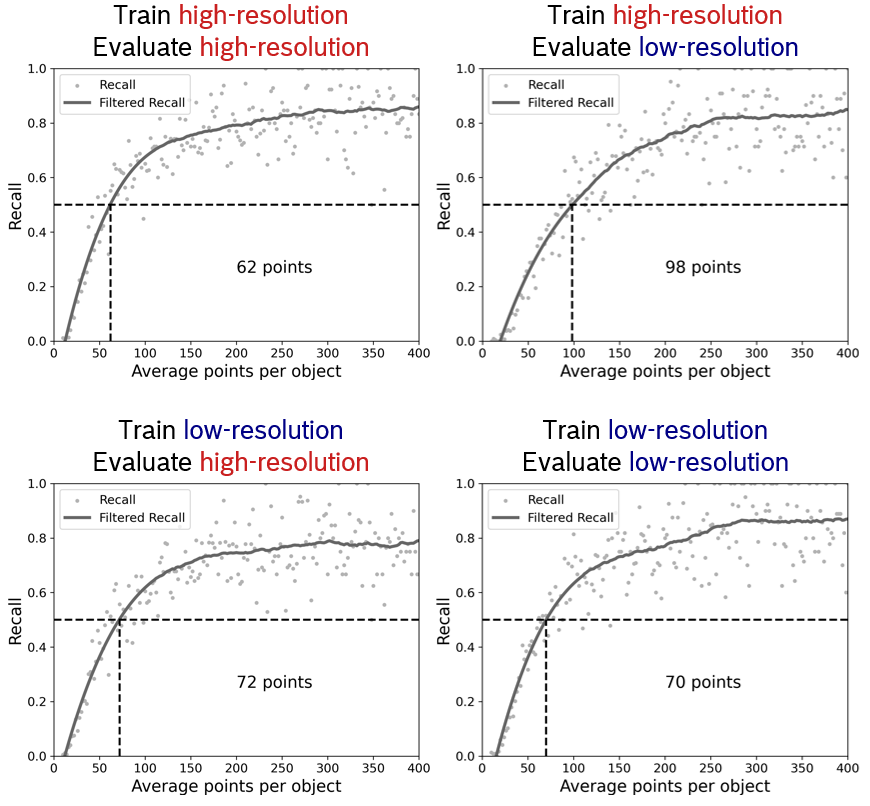}
	\caption{Recall versus average number of points received per object for training and evaluation on all sensor combinations. A recall of 1 means that all real-world objects were detected, while a recall of 0 means that no real-world objects were detected. All plots were filtered with a Savitzky-Golay filter in order to estimate the average number of points received per object required to achieve a recall of 0.5, which is indicated by the dashed black lines.} \label{fig:recall_vs_num_points}
\end{figure}

The relationships between recall and the average number of points received per object for training and evaluation on all sensor combinations is shown in Figure~\ref{fig:recall_vs_num_points}. For all training and evaluation combinations, the recall clearly increases as the average number of points received per object increases. This trend is especially noticable in the region where less than 100 points are received on average per object. In general, this shows that the more points the object detection network receives from an object, the more likely it is that the object detection network is able to distinguish the object from the background.

Each plot also shows the average number of points received per object required to achieve a recall of 0.5 for the given training and evaluation LiDAR sensor combination. These values are summarised in Table~\ref{table:points} for the inference domain gap and the training domain gap.

\begin{table}[H]
	\renewcommand{\arraystretch}{1.3}
	\caption{Average number of points received per object required to achieve a recall of 0.5 for object detection networks trained and evaluated on different combinations of high-resolution and low-resolution LiDAR sensors. Lower is better. Top: Inference domain gap. Bottom: Training domain gap.} \label{table:points} \centering
	\begin{tabular}{|c|c||c|}
		\hline
		Trained on & Evaluated on & Points \\
		\hline
		\hline
		High-res & High-res & \textbf{62} \\
		\hline
		High-res & Low-res & 98 \\
		\hline
		\noalign{\vskip 0.4cm} 
	\end{tabular}
	\begin{tabular}{|c|c||c|}
		\hline
		Trained on & Evaluated on & Points \\
		\hline
		\hline
		Low-res & High-res & 72 \\
		\hline
		Low-res & Low-res & \textbf{70} \\
		\hline
		\noalign{\vskip 0.4cm} 
	\end{tabular}
	\begin{tabular}{|c|c||c|}
		\hline
		Trained on & Evaluated on & Points \\
		\hline
		\hline
		High-res & High-res & \textbf{62} \\
		\hline
		Low-res & High-res & 72 \\
		\hline
	\end{tabular}
	\begin{tabular}{|c|c||c|}
		\hline
		Trained on & Evaluated on & Points \\
		\hline
		\hline
		High-res & Low-res & 98 \\
		\hline
		Low-res & Low-res & \textbf{70} \\
		\hline
	\end{tabular}
\end{table}

\subsubsection{Inference domain gap}
\begin{figure}[H]
	\centering
	\includegraphics[width=12cm]{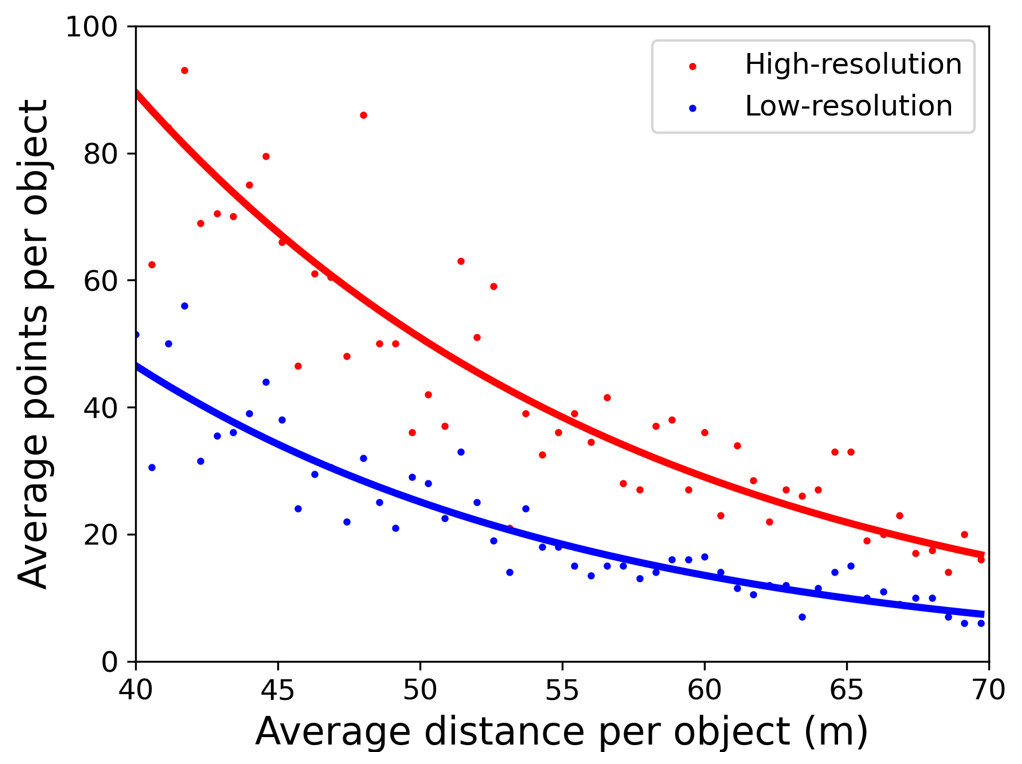}
	\caption{Average number of points received per real-world object versus average distance to the LiDAR sensor per real-world object, fitted with an exponential function. On average, more points are received per object from the high-resolution LiDAR sensor (red) than from the low-resolution LiDAR sensor (blue) and the number of points received per object decreases as the distance to the LiDAR sensor increases.} \label{fig:points_vs_distance}
\end{figure}

At first, the average number of points required to achieve a recall of 0.5 seems to contradict the relative APs in Table~\ref{table:all}. For the model trained on high-resolution LiDAR data, 36 fewer points per object are required on average to achieve a recall of 0.5 when evaluating with the high-resolution instead of the low-resolution LiDAR sensor. However, for the model trained on low-resolution LiDAR data, two additional points per object are required on average to achieve a recall of 0.5 when evaluating with the high-resolution instead of the low-resolution LiDAR sensor, even though the relative AP in Table~\ref{table:all} is higher when evaluating with the high-resolution instead of the low-resolution LiDAR sensor. This indicates that it is not just the average number of points received per object that is important, but also whether these points were recorded with the high-resolution or low-resolution LiDAR sensor. At inference time, the object detection network can detect objects with fewer points when it was trained on data from the same LiDAR sensor. This again suggests that the inference gap can not only be closed, but even reversed by using a high-resolution LiDAR sensor at inference time, because of the additional points received per object. These additional points can be observed by plotting the average number of points received per real-world object versus the average distance of the real-world objects from the LiDAR sensor for the high-resolution and low-resolution LiDAR sensors, as shown in Figure~\ref{fig:points_vs_distance}. This observation explains why the combination of training and evaluating on high-resolution LiDAR data results in the largest performance improvement (relative AP of +25\% shown in Table~\ref{table:all}), because the object detection network requires fewer points on average per object to be able to detect real-world objects and the LiDAR data contains more points on average per object because of the higher-resolution.

\subsubsection{Training domain gap}
For the training domain gap, when evaluating on high-resolution LiDAR data, 10 fewer points per object are required on average when the object detection network is trained on high-resolution rather than low-resolution LiDAR data. When evaluating on low-resolution LiDAR data, 18 fewer points per object are required on average when the object detection network is trained on low-resolution rather than high-resolution LiDAR data. This is consistent with the relative APs in Table~\ref{table:all}, which show that training the object detection network on data from the same LiDAR sensor results in the best performance at inference time. This again indicates that the training gap cannot be closed by increasing the average number of points per object by using a higher-resolution LiDAR sensor at training time. Instead, the performance drops when a different LiDAR sensor is used for training, independent of whether the LiDAR used for training is higher-resolution or lower-resolution than the LiDAR used at inference time.

\section{Conclusions}
In this work, we investigated sensor-to-sensor domain gaps in LiDAR object detection networks. Using a unique dataset, which enabled us to study the domain gaps independent of other effects, we showed that there are actually two distinct domain gaps - the inference domain gap and the training domain gap. The inference domain gap varies strongly with the average number of points per object at inference time and our experiments show that increasing the average number of points per object by using a higher-resolution LiDAR sensor at inference time, can improve the performance of the object detection network, independent of which LiDAR data was used for training. This indicates that it is better to use a high-resolution LiDAR for inference, even if the network has been trained with data from a low-resolution LiDAR. In contrast, the training domain gap does not vary with the number of points per object. Our experiments show that adding more points to the objects at training time, by training with a higher-resolution LiDAR sensor, does not improve performance, but actually worsens the performance. In this case, if inference will be carried out with a low-resolution LiDAR, it is better to train with data from the same low-resolution LiDAR, even if a high-resolution LiDAR is available. These important findings show that different strategies are required to close these distinct inference and training domain gaps.

In the future, the same analysis could be applied to other sensor modalities \cite{FengHaase2020deep}, such as radar \cite{scheiner_radar_2021}. It could also be applied to other domain gaps, such as weather domain gaps, since it is known that significantly fewer LiDAR points are typically received in rainy weather than in dry weather \cite{xu_spg_2021}. In fact, our initial experiments on weather domain gaps indicate very similar behaviour to the sensor-to-sensor domain gaps, since we also observe a distinct inference domain gap which varies strongly with the average number of points per object at inference time and a distinct training domain gap which does not vary with the number of points per object. Again, this highlights the importance of using different strategies to close these distinct inference and training domain gaps.

\section*{Acknowledgements}
The research leading to these results is funded by the German Federal Ministry for Economic Affairs and Energy (BMWi) within the project ``KI Delta Learning" (Förderkennzeichen 19A19013D). The authors would like to thank the consortium for the successful cooperation.

\bibliographystyle{IEEEtran}
\bibliography{references}

\begin{thebibliography}{10}
\providecommand{\url}[1]{#1}
\csname url@samestyle\endcsname
\providecommand{\newblock}{\relax}
\providecommand{\bibinfo}[2]{#2}
\providecommand{\BIBentrySTDinterwordspacing}{\spaceskip=0pt\relax}
\providecommand{\BIBentryALTinterwordstretchfactor}{4}
\providecommand{\BIBentryALTinterwordspacing}{\spaceskip=\fontdimen2\font plus
\BIBentryALTinterwordstretchfactor\fontdimen3\font minus
  \fontdimen4\font\relax}
\providecommand{\BIBforeignlanguage}[2]{{%
\expandafter\ifx\csname l@#1\endcsname\relax
\typeout{** WARNING: IEEEtran.bst: No hyphenation pattern has been}%
\typeout{** loaded for the language `#1'. Using the pattern for}%
\typeout{** the default language instead.}%
\else
\language=\csname l@#1\endcsname
\fi
#2}}
\providecommand{\BIBdecl}{\relax}
\BIBdecl

\bibitem{hnewa_rain_2021}
M.~Hnewa and H.~Radha, ``Object detection under rainy conditions for autonomous
  vehicles: A review of state-of-the-art and emerging techniques,''
  \emph{{IEEE} Signal Processing Magazine}, 2021.

\bibitem{singha_night_2019}
A.~Singha and M.~K. Bhowmik, ``Moving object detection in night time: A
  survey,'' \emph{2nd International Conference on Innovations in Electronics,
  Signal Processing and Communication (IESC)}, 2019.

\bibitem{wang_lidar_2020}
X.~Wang, H.~Pan, K.~Guo, X.~Yang, and S.~Luo, ``The evolution of lidar and its
  application in high precision measurement,'' \emph{IOP Conference Series:
  Earth and Environmental Science}, 2020.

\bibitem{li_data_2019}
J.~Li, Y.~Wong, Q.~Zhao, and M.~S. Kankanhalli, ``Learning to learn from noisy
  labeled data,'' \emph{Proceedings of the {IEEE}/{CVF} Conference on Computer
  Vision and Pattern Recognition (CVPR)}, 2019.

\bibitem{ganin_unsupervised_2015}
Y.~Ganin and V.~Lempitsky, ``Unsupervised domain adaptation by
  backpropagation,'' \emph{Proceedings of the 32nd International Conference on
  International Conference on Machine Learning (ICML)}, 2015.

\bibitem{triess_survey_2021}
L.~Triess, M.~Dreissig, C.~Rist, and J.~M. Z{\"o}llner, ``A survey on deep
  domain adaptation for {LiDAR} perception,'' \emph{Proceedings of the {IEEE}
  Intelligent Vehicles Symposium ({IV}) Workshops}, 2021.

\bibitem{wirges_single-stage_2020}
S.~Wirges, S.~Ding, and C.~Stiller, ``Single-stage object detection from
  top-view grid maps on custom sensor setups,'' \emph{{IEEE} Intelligent
  Vehicles Symposium ({IV})}, 2020.

\bibitem{rist_cross_sensor_2019}
C.~Rist, M.~Enzweiler, and D.~Gavrila, ``Cross-sensor deep domain adaptation
  for lidar detection and segmentation,'' \emph{{IEEE} Intelligent Vehicles
  Symposium ({IV})}, 2019.

\bibitem{qin_multiscale_2019}
C.~Qin, H.~You, L.~Wang, C.-C.~J. Kuo, and Y.~Fu, ``Pointdan: A multi-scale 3d
  domain adaption network for point cloud representation,'' \emph{Conference on
  Neural Information Processing Systems (NeurIPS)}, 2019.

\bibitem{xu_spg_2021}
Q.~Xu, Y.~Zhou, W.~Wang, C.~R. Qi, and D.~Anguelov, ``{SPG}: Unsupervised
  domain adaptation for 3d object detection via semantic point generation,''
  \emph{Proceedings of the {IEEE}/{CVF} International Conference on Computer
  Vision (ICCV)}, 2021.

\bibitem{geiger_kitti_2013}
A.~Geiger, P.~Lenz, C.~Stiller, and R.~Urtasun, ``Vision meets robotics: The
  kitti dataset,'' \emph{International Journal of Robotics Research (IJRR)},
  2013.

\bibitem{caesar_nuscenes_2020}
H.~Caesar, V.~Bankiti, A.~H. Lang, S.~Vora, V.~E. Liong, Q.~Xu, A.~Krishnan,
  Y.~Pan, G.~Baldan, and O.~Beijbom, ``nuscenes: A multimodal dataset for
  autonomous driving,'' \emph{{IEEE}/{CVF} Conference on Computer Vision and
  Pattern Recognition (CVPR)}, 2020.

\bibitem{Yang_2018_CVPR}
B.~Yang, W.~Luo, and R.~Urtasun, ``Pixor: Real-time 3d object detection from
  point clouds,'' \emph{Proceedings of the {IEEE} Conference on Computer Vision
  and Pattern Recognition (CVPR)}, 2018.

\bibitem{FengHaase2020deep}
D.~Feng, C.~Haase-Sch{\"u}tz, L.~Rosenbaum, H.~Hertlein, C.~Glaeser, F.~Timm,
  W.~Wiesbeck, and K.~Dietmayer, ``Deep multi-modal object detection and
  semantic segmentation for autonomous driving: Datasets, methods, and
  challenges,'' \emph{{IEEE} Transactions on Intelligent Transportation
  Systems}, 2020.

\bibitem{scheiner_radar_2021}
N.~Scheiner, F.~Kraus, N.~Appenrodt, J.~Dickmann, and B.~Sick, ``Object
  detection for automotive radar point clouds - a comparison,'' \emph{AI
  Perspectives}, 2021.

\end{thebibliography}

\end{document}